\documentclass[a4paper, 10pt, conference]{ieeeconf}      
\usepackage{FG2026}

\FGfinalcopy 

\IEEEoverridecommandlockouts                              
\usepackage{graphics} 
\usepackage{epsfig} 
\usepackage{mathptmx} 
\usepackage{times} 
\usepackage{amsmath} 
\usepackage{amssymb}  
\usepackage{graphicx}
\usepackage{multirow}
\usepackage{hyperref} 
\usepackage{tabularx} 
\usepackage{array}    
\usepackage[table]{xcolor}
\usepackage{cleveref}
\usepackage{tablefootnote}

\usepackage{booktabs}
\usepackage{makecell}
\usepackage{siunitx}
\def\FGPaperID{254} 

\title{\LARGE \bf
Context Matters: Vision-Based Depression Detection Comparing Classical and Deep Approaches
}

\author{\parbox{16cm}{\centering
    {\large Maneesh Bilalpur$^1$, Saurabh Hinduja$^2$, Sonish Sivarajkumar$^1$, Nicholas Allen$^3$, Yanshan Wang$^{1,4}$, Itir Onal Ertugrul$^5$, and Jeffrey F. Cohn$^{1,6}$}\\
    {\normalsize
    $^1$Intelligent Systems Program, University of Pittsburgh, Pittsburgh, USA\\
    $^2$CGI Technologies and Solutions Inc, USA\\
    $^3$University of Oregon, USA\\
    $^4$Department of Health Information Management, University of Pittsburgh, Pittsburgh, USA\\
    $^5$Utrecht University, the Netherlands\\
    $^6$Department of Psychology, University of Pittsburgh, Pittsburgh, USA\\}}
    \thanks{Email: Maneesh Bilalpur (mab623@pitt.edu)}
}

\begin{document}

\ifFGfinal
\thispagestyle{empty}
\pagestyle{empty}
\else
\author{Anonymous FG2026 submission\\ Paper ID \FGPaperID \\}
\pagestyle{plain}
\fi
\maketitle

\begin{abstract}

The classical approach to detecting depression from vision emphasizes interpretable features, such as facial expression, and classifiers such as the Support Vector Machine (SVM). With the advent of deep learning, there has been a shift in feature representations and classification approaches. Contemporary approaches use learnt features from general-purpose vision models such as VGGNet to train machine learning models. Little is known about how classical and deep approaches compare in depression detection with respect to accuracy, fairness, and generalizability, especially across contexts. To address these questions, we compared classical and deep approaches to the detection of depression in the visual modality in two different contexts: Mother-child interactions in the TPOT database 
and patient-clinician interviews in the Pitt database. 
In the former, depression was operationalized as a history of depression per the DSM and current or recent clinically significant symptoms. In the latter, all participants met initial criteria for depression per DSM, and depression was reassessed over the course of treatment.  The classical approach included handcrafted features with SVM classifiers. Learnt features were turn-level embeddings from the FMAE-IAT that were combined with Multi-Layer Perceptron classifiers.  The classical approach achieved higher accuracy in both contexts. It was also significantly fairer than the deep approach in the patient-clinician context. Cross-context generalizability was modest at best for both approaches, which suggests that depression may be context-specific.
\end{abstract}

\section{INTRODUCTION}
Depression is a recurrent and highly prevalent condition that entails significant personal impairment and suffering, increases risk for suicide, and results in substantial economic costs for the individual and society \cite{burcusa2007risk}. Approximately 20 million cases are reported yearly in the United States. The economic burden of depression exceeds a quarter of a trillion dollars in healthcare and related costs \cite{greenberg2023economic}. Automated approaches that leverage objective measures of human behavior for depression detection offer a reliable solution to limit the socio-economic impact of depression through early intervention and provide an accessible means of assessing depression over time.



We focus on the vision modality to detect depression. Depression relevant channels in the vision modality include facial expression \cite{girard2014nonverbal, scherer2014automatic, song2020spectral} and head motion \cite{kacem2018detecting, daoudi2019gram, dibekliouglu2017dynamic}. The majority of studies in this area are limited to a single database \cite{arioz2022scoping}, lack well-validated clinical diagnosis of depression, fail to assess fairness between subgroups (e.g., race or sex) and give little attention to generalizability across different contexts.

The advent of deep learning has led to a decreased focus on handcrafted features with conventional classifiers and promoted feature learning with classification. Little is known about how deep approaches compare to classical approaches with respect to the accuracy of depression detection, demographic fairness, and generalizability across contexts. Although the fairness of machine learning approaches has received much traction in computer vision problems such as image captioning \cite{zhao2021understanding, hirota2022quantifying}, visual recognition \cite{ranjit2023variation, birhane2024dark}, visual question-answering \cite{zhang2016yin}, and more (see survey \cite{dehdashtian2024fairness}), most depression detection approaches are primarily focused on the detection performance. The evidence for the effect of interaction context and cross-context generalizability in depression detection is little, if any. We compare a classical approach (handcrafted features with a Support Vector Machine classifier) and a deep approach (learnt features and a Multi-Layer Perceptron classifier) to answer the following questions:



\begin{enumerate}
\item How do the deep and the classical approaches compare with respect to the accuracy of depression detection?
\item How does demographic fairness differ between the deep and the classical approaches? 
\item Does one approach generalize across contexts (mother-child interaction vs. patient-clinician interview) better than the other?
\end{enumerate}



We used two very different databases: mother-child interactions \cite{sheeber2023maternal, nelson2012african, nelson2021affective} and patient-clinician interviews \cite{cohn2009detecting, frank2011predictors, girard2014nonverbal, yang2012detecting} with clinically valid depression status. Learnt features from the pretrained FMAE-IAT \cite{ning2024representation} were used to train a MultiLayer Perceptron (MLP) classifier for the deep approach. In comparison, the handcrafted features were developed to capture Face and Head Dynamics (FHD) and Action Units (AU) channels; these were trained with the Support Vector Machine (SVM) classifier. 


Both deep and classical approaches were evaluated in the two diverse interaction contexts for accuracy at depression detection. Demographic fairness was studied to identify disparities in terms of sex and race of individuals. To evaluate cross-context generalizability, the model trained in one context was tested in the unseen context. The results suggest that the classical approach is more accurate the deep approach in depression detection. The fairness of depression detection is sensitive to the interaction context. Both approaches are comparably fair in the mother-child interactions. However, the classical approach was fairer than the deep approach in the patient-clinician interviews. Cross-context generalizability was found to be challenging for both approaches.

\section{RELATED WORK}
\subsection{Classical and Deep Approaches}
Face and head motion dynamics, and action units are some of the well-studied handcrafted visual features in depression. Cohn et al. \cite{cohn2009detecting} was one of the early large-scale studies to find that depression in clinical interviews can be detected using automatic measures of face dynamics. They found that automatic measures of face dynamics achieve close to 80\% accuracy in detecting depression. They also utilized 17 manually coded facial action units to achieve similar accuracy. Interestingly, the lone AU 14 (Buccinator muscle) detected depression with 88\% accuracy. Scherer et al. \cite{scherer2014automatic} found that depressed individuals smiled often but briefly and Girard et al. \cite{girard2014nonverbal} found that non-verbal behaviors can capture severity-sensitive manifestations associated with depression. They found that depressed subjects expressed fewer affiliative expressions (such as smiles) but more non-affiliative expressions (such as contempt). They also found reduced head motion with depression. 

Daoudi et al. \cite{daoudi2019gram} extracted kinematics (velocity and acceleration) of body shape trajectories for depression severity prediction. Changes in body pose were represented as Bezier curves that were used for extracting the kinematics of pose changes. These were encoded on the video-level and depression severity was predicted using SVM. Until Kacem et al. \cite{kacem2018detecting}, existing work on face and head dynamics determined the velocity and acceleration dynamics of face and head motion using Euclidean geometry methods on a non-linear manifold. Their work projected facial landmarks into a linear manifold space using a barycentric coordinate representation of landmarks where Euclidean geometry constraints hold. This resulted in improved depression severity prediction accuracy using facial dynamics and demonstrated that their representation offered an interpretable approach to quantifying the psychomotor retardation in depression. 

In deep learning based approaches, Al et al. \cite{al2018video} fine-tuned pretrained 3D CNN model over short clips from longer videos and then accounted for the temporal nature of the videos using an RNN network. Their two-stream video model used both aligned and non-aligned faces. The two-stream CNN-RNN outputs were mean-pooled for the final BDI score prediction. Song et al. \cite{song2020spectral} represented the video-level AU, headpose, and gaze features from the DAIC-WoZ interviews as magnitude and phase spectra to handle inputs of variable duration. They compared the performance of 1D-CNN trained with a magnitude-phase representation of the multimodal features against a simple MLP trained with the vectorized representation of the magnitude and phase components and found that the vectorized representation of modalities is efficient for depression severity prediction. 

\subsection{Fairness and Generalizability}

Fairness for gender as a sensitive attribute is being increasingly studied \cite{cheong2024fairrefuse, cheong2023towards} in depression detection. \cite{cheong2023towards} comprehensively studied gender bias for its sources and evaluated different mitigation methods in multimodal approaches. To our knowledge, only \cite{cheong2024fairrefuse} has studied unimodal vision for fairness in depression detection. They proposed mitigating gender bias by preventing the classifier from implicitly relying on gender-specific information in the features by marginalizing them. 


Limited literature exists on cross-context generalizability of depression detection approaches \cite{alghowinem2020interpretation, alghowinem2015cross}. \cite{alghowinem2015cross} studied the generalizability of head dynamics and eye activity by training on one dataset and testing on a different, unseen dataset. They found that testing on unseen datasets can lead to significantly attenuated performance. Alternatively, a recent work \cite{alghowinem2020interpretation} studied feature generalizability by selecting optimal features on a source dataset(s) to use them independently for training and testing on the unseen target dataset. Compared to the model generalizability, feature generalizability was found to be more effective. 


In addition to the limited literature on fairness and cross-context generalizability, most of the existing work on depression detection is limited to a single database \cite{arioz2022scoping} of human-agent interviews \cite{gratch2014distress, ringeval2019avec}. This relative lack of context diversity limited the understanding of context and depression detection. To address these limitations, this work studies depression detection in two diverse contexts with clinically valid depression status. In the two contexts, the deep approach was compared to the classical approach for performance at depression detection, fairness by demographic subgroups, and cross-context generalizability.


\section{DATASETS}

The mother-child interaction context used in this work is a part of the Transitions in Parenting of Teens (TPOT) database \cite{sheeber2023maternal, nelson2012african, nelson2021affective}. It includes a depressed group, i.e., mothers with a history of treatment for depression and current or recent elevation in depressive symptoms (PHQ-8 $>$ 10) \cite{kroenke2009phq}; and a non-depressed group, i.e., mothers with no lifetime history of treatment for depression and no more than mild depressive symptoms currently (PHQ-8 $<$ 8). The time between the assessment of current depression and the mother-child interaction varied from same day to several weeks. The mother-child interaction consisted of a problem-solving task in which mother and child were asked to identify and resolve a problem from an updated Issues Checklist \cite{prinz1979multivariate}. This problem-solving interaction lasted about 15-minutes. 84\% of participants were White; minorities were distributed across American Indian/Alaskan, Native Hawaiian/Pacific Islander, African American, or multiple ethnicities. The age range in children was limited to early adolescence, and all families were low-income. Audio was captured at 16kHz and video at 30fps and 720p resolution with dedicated hardware for mother and child. 148 dyads, of which 73 dyads were in the depressed group.


The patient-clinician interview context used in this work is a part of the University of Pittsburgh depression database (Pitt) \cite{cohn2009detecting, frank2011predictors, girard2014nonverbal, yang2012detecting}. Patients undergoing treatment for clinically diagnosed depression were interviewed by clinicians at regular intervals to assess their depression severity on the Hamilton Rating Scale for Depression (HRSD) \cite{hamilton1960rating} scale. Fifty-seven depressed participants were interviewed for depression severity up to four occasions at 1, 7, 13, and 21 weeks post diagnosis by clinical interviewers (all female). Depression was defined as an HRSD score of 15 or higher \cite{cohn2009detecting}. Three cameras captured the patient, and a dedicated camera recorded the interviewer. Audio was recorded using dedicated lavalier microphones for the patient and the interviewer. Only the video from the frontal-facing camera of the patient was used. Fifty participants with valid audio-visual recordings and transcriptions are used in this work. For consistency with existing work \cite{dibekliouglu2017dynamic, kacem2018detecting, alghowinem2020interpretation}, all 135 sessions from the fifty participants were treated as independent observations for depression detection. 


The segmentation of utterances and their transcription was performed manually for both databases. Each utterance was identified by its speaker along with the start and stop timestamps. For comparability between databases, the focal person for depression detection was limited to mothers in TPOT and patients in Pitt. The race and sex distributions are presented as  
\autoref{tab:race_distribution}. Note that sessions with unknown race information were excluded from the fairness analysis. \autoref{tab:dataset_stat} presents statistics for sessions, utterances, and turns for depression detection in both databases. 



\begin{table}[h!]
  \centering
  \caption{Participant demographics. Numbers in parentheses are the sessions\tablefootnote{Unlike the mother-child dataset, the participants in the patient-clinician dataset were interviewed at regular intervals during their treatment. Hence, the participants:sessions ratio was variable.}}
    \begin{tabular}{cccc}
    \hline
    \textbf{Demographic} & \textbf{Subgroup}  & \textbf{Mother-Child}  & \textbf{Patient-Clinician} \\
    \hline
    & White & 126   & 43 (119) \\
    Race & Non-white\tablefootnote{Non-white included Native American, Hawaiian American, Black, Asian, and multiple races.} & 20    & 7 (16) \\
    & Unknown   & 2     & - \\
    \hline
    & Men & -     & 18 (48) \\
    Sex & Women & 148     & 32 (87) \\
    \hline
    & Total & 148   & 50 \\
    \hline
    \end{tabular}%
    
  \label{tab:race_distribution}%
\end{table}%



\begin{table}[htbp]
  \centering
  \caption{Dataset statistics}
  \resizebox{\linewidth}{!}{
    \begin{tabular}{ccccc}
    \hline
          &       & \textbf{Total} & \textbf{Depressed} & \textbf{Non-depressed} \\
    \hline
          & \# sessions & 148   & 73    & 75 \\
    \textbf{Mother-Child} & \# utterances & 34,072 & 16,354 & 17,718 \\
          & \# turns & 18,346 & 8,446  & 9,900 \\
    \hline
          & \# sessions & 135   & 81    & 54 \\
    \textbf{Patient-Clinician} & \# utterances & 16,302 & 10,759 & 5,543 \\
          & \# turns & 12,363 & 8,031  & 4,332 \\
    \hline
    \end{tabular}%
    }
  \label{tab:dataset_stat}
\end{table}%

\section{METHODS}
This subsection describes features used, classifier configurations, and the training framework for training our deep and classical approaches.

\subsection{Deep Approach}


\begin{figure}[h!]
    \centering
    \includegraphics[width=\linewidth]{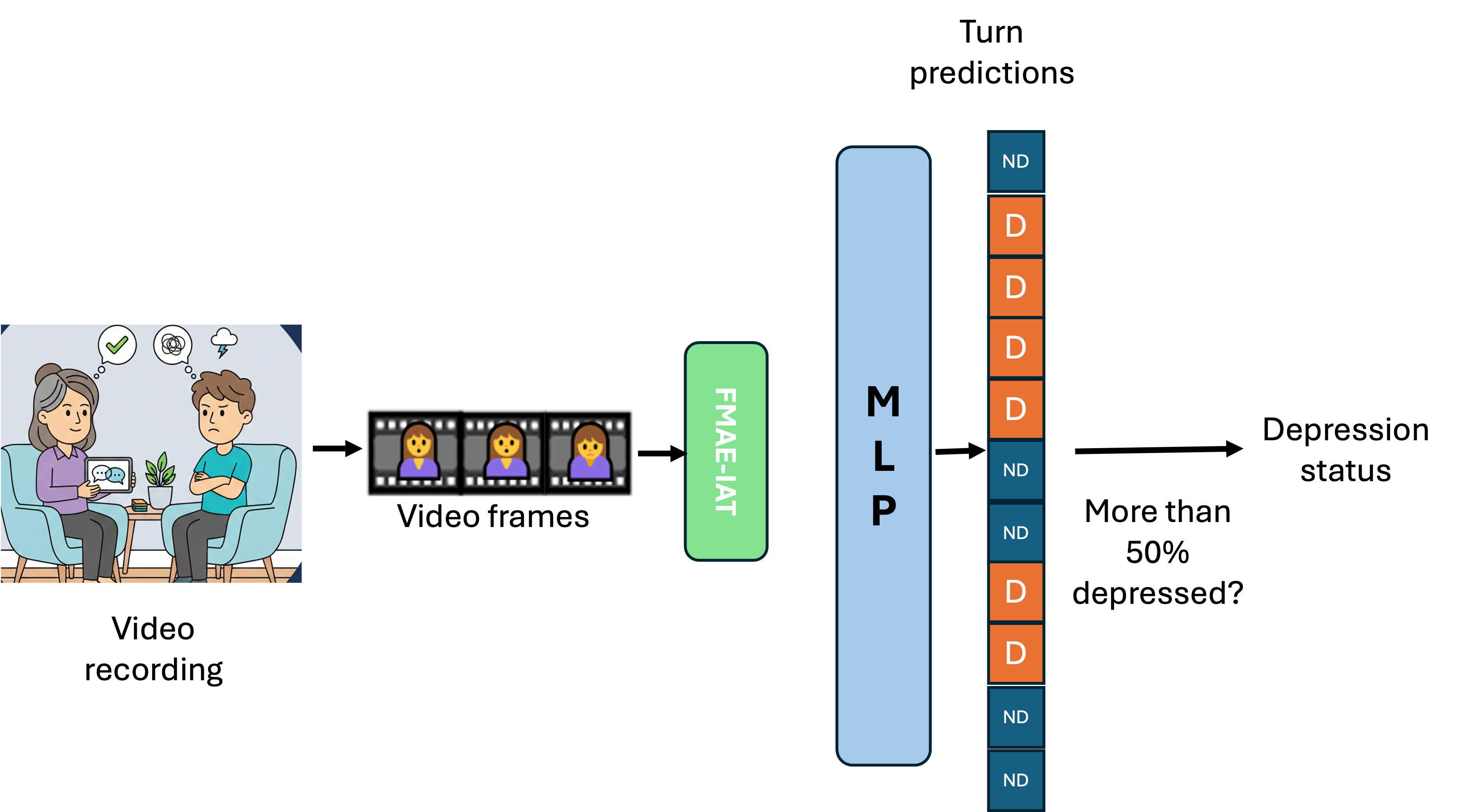}
    \caption{Overview of the deep approach using turn-level embeddings from the FMAE-IAT as features for the MLP classifier. Session-level predictions were aggregated using a majority-voting criterion.}
    \label{fig:deep_approach_overview}
\end{figure}

\begin{figure}[h!]
    \centering
    \includegraphics[width=\linewidth]{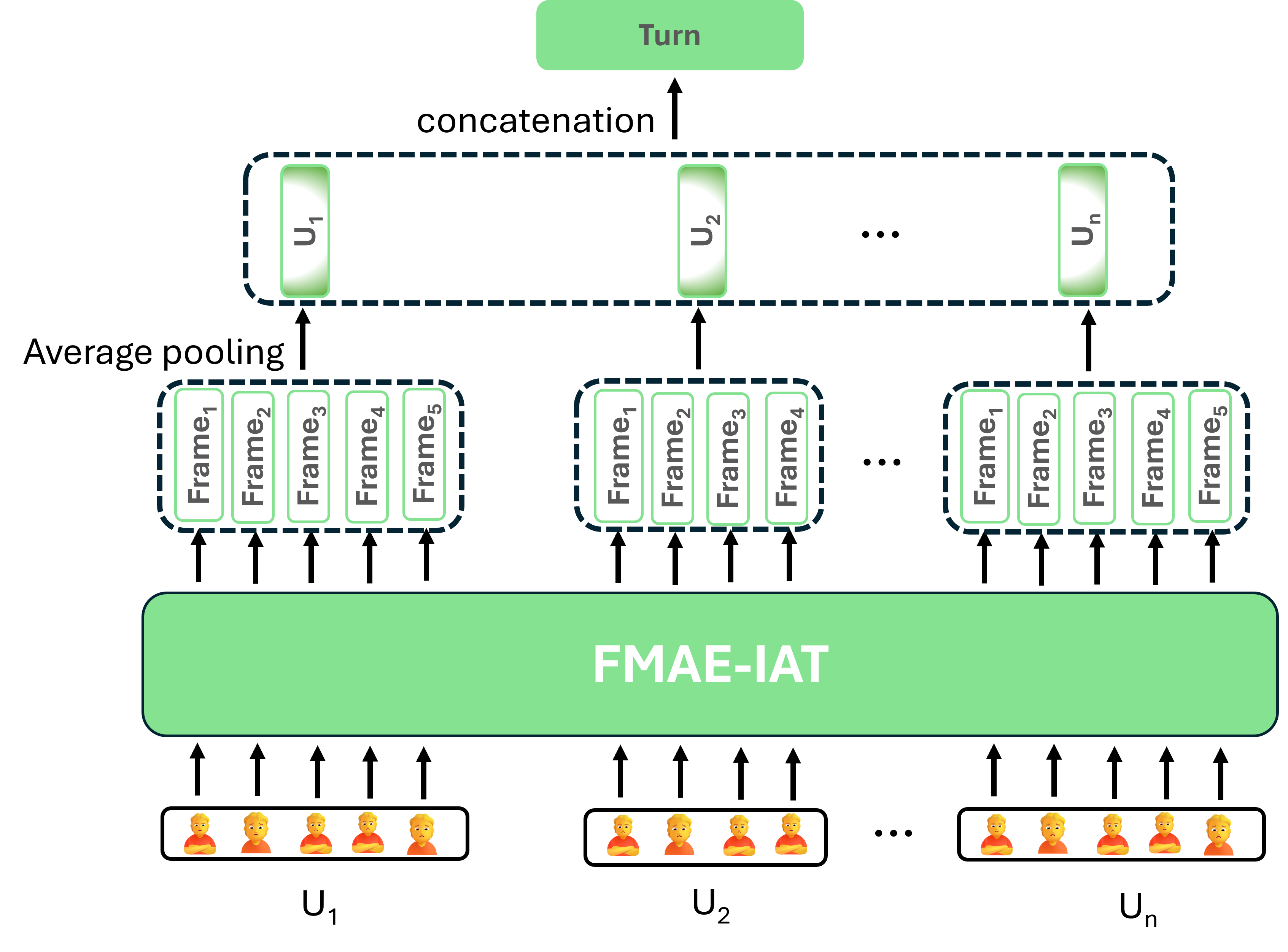}
    \caption[Turn representation for vision modality]{Turn representation: FMAE-IAT embeddings for all frames within an utterance are extracted. Average-pooling is performed to represent an utterance ($U_{i}$). Concatenation of utterances within a turn is used as the vision modality representation.}
    \label{fig:vision_turn_representation}
    \label{fig:placeholder}
\end{figure}

\begin{table*}[th!]
  \centering
  \caption{Overview of contextual differences between the TPOT and Pitt databases}
  
  \rowcolors{3}{gray!10}{white}
  
    \begin{tabularx}{\textwidth}{@{} >{\arraybackslash}p{3.5cm} >{\arraybackslash}X >{\arraybackslash}X @{}}
    \hline
    \textbf{Database characteristics} & \textbf{TPOT} & \textbf{Pitt} \\
    \hline
    Focal person & Mothers & Clinical trial participants \\
    Participant sex & Women-only & Women and men \\
    Task  & Unstructured problem-solving & Depression severity assessment \\
    Depression definition & History of depression and current symptoms & Current depression \\
    Assessment \& interaction lag & Variable lag between depression assessment and behavior sample & No lag \\
    Recordings & Off-line-of-sight camera & Near line-of-sight camera \\
    \hline
    \end{tabularx}%
  \label{tab:dataset_differences_in_crossdomain_generalization}%
\end{table*}%

  
  

The vision modality representation for the deep approach uses embeddings from the FMAE-IAT model (overview \autoref{fig:deep_approach_overview}). The FMAE-IAT was pretrained on a 9M face-image dataset using a masked autoencoding objective, where the model learns to reconstruct the original image from its occluded version. The autoencoding model was then fine-tuned for AU detection using an identity-adversarial training strategy. Given the importance of face and head pose features for depression detection, FMAE-IAT is an appropriate feature representation over generic natural image-based models like ViT-MAE \cite{he2022masked}. Embeddings from the FMAE-base model\footnote{https://github.com/forever208/FMAE-IAT} were extracted for the face videos of subjects from the TPOT and Pitt databases. Given an interaction, the deep approach was operationalized by predicting depression status for its constituent turns. The turn-level predictions were aggregated for session-level prediction using a 50\% thresholding criterion (i.e., a session was detected to be depressed if at least 50\% of the constituent turns are depressed). Individual turns were defined using utterance-averaged representations (\autoref{fig:placeholder}), i.e., all frames within the constituent utterances were average-pooled and concatenated for the turn representation. Training with turn-level inputs reduces the risk of overfitting in depression detection.



Recent research in large-scale pretrained models suggests that the choice of embedding layer influences downstream task performance. This was observed across a variety of tasks such as image retrieval and segmentation \cite{walmer2023teaching}, and synthetic image detection \cite{koutlis2024leveraging}. More importantly, this has also been observed in depression detection (e.g., \cite{wu2023self} suggests early layers are useful for emotion detection while later layers are useful for depression detection), however, with audio modality. In order to to validate the layer-sensitivity of vision embeddings for depression detection and identify the optimal embeddings from the FMAE-IAT model, the standard linear probing approach was used. In linear probing, a logistic regression (LR) was trained to detect depression using utterance-level embeddings from each layer of the pretrained model. To avoid peeking ahead into the test set, the optimal layer was selected based on the performance of the classifier on the validation set.




A simple two-layer MLP classifier was used for the classifier in the deep approach. Adam optimizer \cite{adam2014method} together with class weighting in the cross-entropy loss function were used to account for the class imbalance. 

\subsection{Classical Approach}
The classical approach uses handcrafted features to capture face and head dynamics, and action units. Video recordings were used to extract head pose, facial landmarks, and AUs using the AFAR toolbox \cite{ertugrul2019afar}. The resultant features were used to capture summary statistics of intensity, likelihood and duration of occurrence for action units. Similarly for face and head dynamics, the summary statistics for displacement, velocity and acceleration along roll, pitch and yaw, duration of head orientations (looking left/right, up/down, clockwise/anti-clockwise), rate of change in head orientation, displacement, velocity and acceleration for 49 facial landmarks, distance between eyelids, eye-closure duration, and blink rate were used. The summary statistics used were mean, minimum, maximum, standard deviation, variance and interquartile range. The formulations for these handcrafted features follow the convention from existing works on depression detection \cite{alghowinem2013eye, alghowinem2020interpretation, bilalpurmultimodal}. This resulted in 156 features for action units, and 137 features for face and head dynamics. These features were used with an SVM classifier for the handcrafted approach.

\subsection{Fairness}
Benchmarking the fairness of prediction models is necessary to evaluate and mitigate the models from learning unintended biases. It is of extreme importance in healthcare applications due to the catastrophic implications of a biased classifier. We define a fair classifier as one that justly classifies depressed and non-depressed classes without overpredicting or underpredicting a particular demographic group. To quantify the fairness of the classifiers, the Equalized Odds Ratio (EOR) metric \cite{hardt2016equality} was used. The EOR\footnote{https://fairlearn.org/} accounts for bias in both true positive and false positive rates across demographic groups in the prediction. It ranges between 0 (unfair) and 1 (most fair), and \autoref{eq:equalized_odds_ratio} is the mathematical formula for the EOR. This study analyses fairness for sex and race demographics. The fairness analysis for sex included men and women in Pitt (note that the mother-child dataset lacks sex diversity). However, fairness for race was studied by comparing white and non-white groups, given the distributional skew (see \autoref{tab:race_distribution}).

\begin{equation} \label{eq:equalized_odds_ratio}
\begin{aligned}
\text{EOR} &= \min(\text{TPR ratio, FPR ratio}) \\[1em] 
\text{TPR ratio} &= \frac{\min(\text{TPR}_{\text{white}}, \text{TPR}_{\text{non-white}})}{\max(\text{TPR}_{\text{white}}, \text{TPR}_{\text{non-white}})} \\
\text{FPR ratio} &= \frac{\min(\text{FPR}_{\text{white}}, \text{FPR}_{\text{non-white}})}{\max(\text{FPR}_{\text{white}}, \text{FPR}_{\text{non-white}})}\\
\end{aligned}
\end{equation}
where $\text{TPR}_{\text{white}}$ is the True Positive rate for the white population, and $\text{FPR}_{\text{white}}$ is the False Positive Rate for the white population.

\subsection{Cross-context Generalizability}

The contextual differences between the TPOT and Pitt databases are highly varied. Beyond the focal person, they differ in terms of the task, depression definition, assessment approach, etc. These differences as summarized in \autoref{tab:dataset_differences_in_crossdomain_generalization} provide an interesting opportunity to study the cross-context generalizability. 

The cross-context generalizability framework includes training and validating on one context to test on the unseen context. Existing literature in generalizability \cite{alghowinem2015cross, alghowinem2020interpretation} suggests feature generalizability to be more effective than model generalizability in unseen datasets. However, since learnt features in the deep approach lack a feature selection framework, the cross-context generalizability in this work follows the model generalizability framework used in \cite{alghowinem2015cross}.

\subsection{Cross-validation and Metrics}
The cross-validation setup includes a subject-independent 5-fold cross-validation to prevent the models from learning identity-related features. The handcrafted features are normalized using the min-max normalization determined on the training set. The SVM hyperparameters for the classical approach are chosen through a grid search for the choice of kernel and cost of misclassification. Due to the superior performance of early fusion over late fusion in our early experiments, the AU + FHD approach refers to the early fusion of AU and FHD features.

As noted previously, to adjust for the class imbalance, balanced accuracy (also called average recall) was reported along with Positive Agreement (PA), and Negative Agreement (NA) metrics. Note that PA is equivalent to the F1-score in a two-class classification problem. Formulae for PA and NA can be found in \cite{girard2017sayette}. For notational convenience, balanced accuracy is referred to as accuracy (ACC). 

\section{RESULTS}
\subsection{Choice of Embedding Layer}


\begin{table}[!h]
  \centering
  \caption{Validation set accuracy to determine the optimal choice of embedding layer. Numbers in bold are the best embeddings.}
\begin{tabular}{c|cc}
\hline
\makecell{\textbf{Embedding}\\\textbf{layer}} &
\makecell{\textbf{Mother-Child}} &
\makecell{\textbf{Patient-Clinician}} \\
\hline
1 & 0.608 & 0.571 \\
\textbf{2} & 0.649 & \textbf{0.605} \\
3 & 0.611 & 0.600 \\
4 & 0.621 & 0.588 \\
5 & 0.673 & 0.591 \\
\textbf{6} & \textbf{0.683} & 0.576 \\
7 & 0.639 & 0.552 \\
8 & 0.606 & 0.500 \\
\hline
\end{tabular}
\label{tab:optimal_embeding_layer}
\end{table}%

The results (\autoref{tab:optimal_embeding_layer}) for the choice of embedding layer experiment suggest an observable sensitivity of depression detection performance by the embedding layer. Upto a 10\% difference in the validation set performance could be observed between the most optimal and least optimal embedding layers in both the mother-child interaction and the patient-clinician interview contexts. In addition, the optimal embedding layer differed by the context. In mother-child interactions, the sixth layer of FMAE-IAT was optimal for depression detection, while in patient-clinician interviews, the second layer was optimal. For the remainder of the experiments, these optimal embeddings were used as features of the deep approach.

\subsection{Detection Performance of Deep and Classical Approaches}

\begin{table}[h!]
  \centering
  \caption{Performance comparison of deep to classical approach. * indicates a significant difference.}
    \begin{tabular}{cc|ccc}
    \hline
    \textbf{Context} & \textbf{Features} & \textbf{ACC} & \textbf{PA}    & \textbf{NA} \\
    \hline
    \multirow{4}[2]{*}{\begin{tabular}[c]{@{}c@{}}Mother-Child \\ \end{tabular}} & Deep  & 0.597 & 0.607 & 0.514 \\
          & AU    & 0.603 & 0.618 & 0.560 \\
          & FHD   & 0.470  & 0.366 & 0.525 \\
          & AU + FHD & \textbf{0.623} & \textbf{0.635} & \textbf{0.601} \\
    \hline
    \multirow{4}[2]{*}{\begin{tabular}[c]{@{}c@{}}Patient-Clinician \\ \end{tabular}} & Deep  & 0.583 & 0.448 & 0.532 \\
          & AU & 0.621 & \textbf{0.632} & 0.569 \\
          & FHD & 0.625 & 0.552 & \textbf{0.671} \\
          & AU + FHD & \textbf{0.634*} & 0.623 & 0.630 \\
    \hline
    \end{tabular}%
  \label{tab:compare_vision_deep_and_classical_acc}%
\end{table}%

The comparison of deep to classical approaches is presented in \autoref{tab:compare_vision_deep_and_classical_acc}. In mother-child interactions, the deep approach performed with 0.597 accuracy. However, when the class-specific agreements were compared, the agreement for the depressed class (PA=0.607) was found to be higher than the non-depressed class (NA=0.517). Similarly, in the classical approach, the handcrafted AU features can detect depression with 0.603 accuracy, 0.618 PA, and 0.560 NA. The FHD performed poorly with 0.470 accuracy, 0.366 PA, and 0.525 NA. The poor performance of the FHD is similar to existing observations \cite{bilalpur2023shap} where FHD performed poorly as compared to other handcrafted features, including prosody, speech behavior, and linguistic features. The fusion of AU and FHD (AU + FHD) improved the detection performance to 0.623 accuracy, along with class-specific agreements of 0.635 PA and 0.601 NA. When the deep approach was compared to its equivalent classical approach (i.e., AU + FHD), it was noticeable that the deep approach underperformed the classical approach. A McNemar`s test\footnote{https://statsmodels.org} comparing the two approaches (\textit{Deep} to the \textit{AU+FHD}) revealed a significant difference ($p<0.05$) between the predictions of the two approaches. This significance was not observed in the mother-child interaction context ($p>>0.1$). Also importantly, the high discrepancy between PA and NA in the deep approach suggests that it is better at predicting only the depressed class. This discrepancy was not observed with the classical approach. 

Similar observations were made in patient-clinician interviews. The deep approach for depression detection resulted in 0.583 accuracy, 0.448 PA, and 0.532 NA. The classical approach with AU features performed better than the deep approach with 0.621 accuracy, 0.632 PA, and 0.569 NA. Unlike the mother-child interactions, the FHD features in patient-clinician interviews detected depression with better than chance-level performance with 0.625 accuracy, 0.552 PA, and 0.671 NA. The fusion of AU and FHD resulted in an improved performance with 0.634 accuracy, 0.623 PA, and 0.630 NA. When the deep and the classical approaches were compared, the classical approach (AU + FHD) features performed better than the deep approach. Although a high discrepancy was observed in class-specific agreement with the deep approach, unlike mother-child interactions, the deep approach in the patient-clinician interviews was better at detecting the depressed class. No such discrepancy was observed in the classical approach with the fusion of AU and FHD features.  




\subsection{Fairness of Deep and Classical Approaches}

\begin{table}[h!]
  \caption{Fairness comparison of deep to classical approach. Table presents the Equalized Odds Ratio (EOR) for race and Sex. EOR=1 implies most fair classifier.}
  \centering
    \begin{tabular}{cc|cc}
    \hline
    \textbf{Context} & \textbf{Features} & \textbf{Race} & \textbf{Sex} \\
    \hline
    \multirow{4}[2]{*}{\begin{tabular}[c]{@{}c@{}}Mother-Child \\ \end{tabular}} & Deep  & \textbf{0.727} & N/A \\
          & AU    & 0.646 & N/A \\
          & FHD   & 0.686 & N/A \\
          & AU + FHD & 0.721 & N/A \\
    \hline
    \multirow{4}[2]{*}{\begin{tabular}[c]{@{}c@{}}Patient-Clinician \\ \end{tabular}} & Deep  & 0.541 & 0.741 \\
          & AU & \textbf{0.873} & \textbf{0.979} \\
          & FHD & 0.545 & 0.796 \\
          & AU + FHD & 0.682 & 0.802 \\
    \hline
    \end{tabular}%
  \label{tab:compare_vision_deep_and_classical_fairness}%
\end{table}%

The EOR for fairness for race in the mother-child interactions is presented in \autoref{tab:compare_vision_deep_and_classical_fairness}. The deep approach achieved a fairness of 0.727 EOR for race (white vs. non-white) in mother-child interactions. As compared to the classical approaches with AU, the FHD features were relatively less fair with EORs of 0.646 and 0.686, respectively. The classical approach with fusion of AU and FHD features improved the fairness to 0.721. When the deep and its equivalent classical approach (AU + FHD) were compared, no significant difference was observed for fairness race in mother-child interactions. In patient-clinician interviews, the EOR for the deep approach for race was 0.541. The classical approach with AU features achieved the highest fairness of 0.873, and FHD features had a lower fairness of 0.545. The classical approach (AU + FHD) improved the fairness as compared to the FHD; however, it substantially underperformed the AU features. When the deep and its equivalent classical approaches (AU + FHD) were compared, the deep approach was found to be less fair than the classical approach.

The fairness analysis for sex also resulted in similar observations. The deep approach has a fairness EOR=0.741. The classical approach with AU features is the fairest with 0.979 EOR. The classical approach with FHD is less fair than the AU features, and their fusion resulted in a decrease in the fairness. The comparison of deep and its equivalent classical approach suggests that the classical approach is fairer than the deep approach. 

\subsection{Context-context Generalizability of Deep and Classical Approaches}

\begin{table}[h!]
  \centering
  \caption{Cross-context generalizability comparison of deep to classical approach. A $\rightarrow$ B indicates model trained in context A and tested in context B.}
    \begin{tabular}{cc|ccc}
    \hline
    \textbf{Train $\rightarrow$ Test} & \textbf{Features} & \textbf{ACC} & \textbf{PA}    & \textbf{NA} \\
    \hline
    \multirow{4}[2]{*}{\begin{tabular}[c]{@{}c@{}}Mother-Child $\rightarrow$ \\ Patient-Clinician\end{tabular}} & Deep  & 0.452 & \textbf{0.470} & 0.226 \\
          & AU    & \textbf{0.536} & 0.166 & \textbf{0.689} \\
          & FHD   & 0.500  & 0.097 & 0.522 \\
          & AU + FHD & 0.500 & 0.097 & 0.000 \\
    \hline
    \multirow{4}[2]{*}{\begin{tabular}[c]{@{}c@{}}Patient-Clinician $\rightarrow$\\ Mother-Child\end{tabular}} & Deep  & 0.477 & 0.486 & 0.231 \\
          & AU & \textbf{0.544} & 0.381 & \textbf{0.559} \\
          & FHD & 0.500 & \textbf{0.660} & 0.000 \\
          & AU + FHD & 0.505 & 0.492 & 0.343 \\
    \hline
    \end{tabular}%
  \label{tab:compare_vision_deep_and_classical_generalizability}%
\end{table}%

The cross-context generalizability results for depression detection between mother-child interactions and patient-clinician interviews found that both deep and classical approaches had limited performance (\autoref{tab:compare_vision_deep_and_classical_generalizability}). In mother-child interactions, only the classical approach with AU features had better than chance-performance (0.536 accuracy) as compared to the deep approach or the alternative classical approaches. The high discrepancy between the PA and NA of the classical approach with AU features suggests that, despite the better than chance-performance, its generalizability is limited to the non-depressed class. Similarly, in the patient-clinician interviews, the classical approach with AU features is better generalizable (0.544 accuracy) as compared to the deep approach or alternative classical approaches. Although the PA and NA discrepancy is lower, the model trained on patient-clinician interviews is better at detecting the non-depressed class in mother-child interactions when compared to the depressed class. Interestingly, although the AU and FHD classical approaches were found to generalize better than the deep approach, their fusion did not improve the cross-context generalizability.
\section{DISCUSSION}
The primary goal of this work is to compare deep and classical approaches to depression detection in two interaction contexts. The preliminary experiment on layer-sensitivity of deep features to depression detection revealed that the sixth layer was optimal in the mother-child context, and the second layer was optimal in the patient-clinician context. This suggests that the conventional approach of using final layer embeddings from pretrained models for downstream tasks may not be optimal in depression detection. This observation is also similar to the layer-sensitivity observed with the audio modality \cite{wu2023self} in depression detection, as well as conventional computer vision-based tasks \cite{walmer2023teaching, koutlis2024leveraging}. 

The comparison of depression detection performance revealed that the classical approach (ACC=0.634) outperforms the deep approach (ACC=0.583) in the patient-clinician interview context with statistical significance. However, in the mother-child context, the classical approach (ACC=0.623) is only nominally better than the deep approach (ACC=0.597). Beyond accuracy, a notable observation is the discrepancy between the class agreements. In both contexts, the deep approach was found to have a high discrepancy between the agreements for the depressed and non-depressed classes. This suggests that the deep approach is better at detecting only one of the two classes (i.e., the depressed class in the mother-child interactions and the non-depressed class in patient-clinician interviews). However, no such discrepancy was observed with the classical approach. 

The comparison of fairness resulted in some interesting observations. The deep approach was the fairest ($EOR_{race}$=0.727) for race in the mother-child interaction context. However, the fusion of the classical approach (AU + FHD) ($EOR_{race}$=0.721) was only marginally behind it, suggesting that both approaches are similar in terms of fairness in the mother-child context. In the patient-clinician interview context, the classical approach with AU features was the fairest for race ($EOR_{race}$=0.873) and sex ($EOR_{race}$=0.979) variables. The classical approach (AU + FHD) resulted in improved fairness ($EOR_{race}$=0.682, $EOR_{sex}$=0.802) as compared to the FHD features ($EOR_{race}$=0.545, $EOR_{sex}$=0.796); it still underperformed the classical approach with the AU features. These results suggest that the fairness of depression detection approaches is sensitive to the context. In particular for race, the deep approach was the fairest in the mother-child context, while the classical approach with AU features was the fairest in the patient-clinician context.

The databases used in this work have highly varied contexts (see \autoref{tab:dataset_differences_in_crossdomain_generalization}). They differ in terms of the nature of the interaction, task, depression definition, assessment approach, and demographic distributions. Their cumulative impact posed a significant challenge for generalizability of both deep and classical approaches.  Although nominally better performance was observed with the classical approach with AU features, the low PA suggests that AU features are better at capturing the non-depressed behavior than the depressed behavior. This trend is similar to existing work on the generalizability of the classical approach \cite{alghowinem2015cross}. They found that head motion and eye activity features had lower generalizability across the popular datasets spanning human-machine interactions and clinical interviews from different cultures. Interestingly, the deep approach underperformed the classical approach in cross-context generalizability. Overall, the poor cross-context generalizability of depression detection approaches highlights an opportunity for a new direction in depression detection. 

This work compared a deep and a classical approach to depression detection in two diverse interaction contexts. Some important takeaways include: 
\begin{enumerate}
    \item \textit{Classical approach outperforms the deep approach.} Our findings suggest that the domain knowledge (vis-a-vis handcrafted features) together with simple classifiers like the SVM may outperform the deep approaches with learnt features from large-scale pretrained models and MLP classifiers. Considering the trend of greater focus on deep learning solutions in depression detection \cite{arioz2022scoping, fu2025first, ilias2024cross}, future research should consider pursuing the classical approach as a baseline, wherever possible.

    \item \textit{Context affects the detection accuracy.} In the patient-clinician interview context, the classical approach significantly outperformed the deep approach. However, the difference was only nominal in the mother-child interview context. This suggests that depression detection should be studied in a wider range of contexts. The effect of context is also observable in terms of channel efficacy in depression detection. The FHD features were effective for depression detection in patient-clinician interviews; however, their efficacy decreases in the mother-child interviews. Suggesting that the channel utility is also dependent on the context. 

    \item \textit{Fairness is context dependent.} In terms of demographic fairness for race and sex, the deep approach was similar to the classical approach in the mother-child interaction context. However, the classical approach was significantly fairer than the deep approach in the patient-clinician interview context. This suggests that the classical approach is at least as fair as the deep approach.

    \item \textit{Cross-context generalizability is a challenge.} The limited cross-context generalizability of deep and classical approaches 
    suggests that depressive behavior may in part be context-dependent. Whether context-agnostic representations of depression are possible remains an open question.
\end{enumerate}

\section{CONCLUSION}

We compared two approaches that use visual features to detect depression. Comparisons included accuracy, fairness, and generalizability in two very different contexts, mother-child interaction and patient-clinician interviews. The classical approach achieved greater accuracy in both contexts.  Results for fairness were mixed. While both approaches were comparable in the patient-clinician context, fairness was greater for the classical approach in the mother-child context. Neither approach generalized well between the two contexts. A likely reason may be that depression manifests differently in a structured clinical interview than in a relatively unstructured interaction between a mother and her child.  

To the best of our knowledge, this is the first vision-centric work to evaluate the efficacy of a deep and a classical approach to depression detection for accuracy, fairness, and generalizability across diverse interaction contexts. This is also one of relatively few studies that use well-validated diagnosis of depression.  With few exceptions, prior work has relied on self-reported depression, which may differ from clinical diagnosis. 

Several limitations may be noted. First, while depression was considered in two very different contexts, many other ecologically important contexts, such as work and home, might be considered. Second, only low-income women were included in the mother-child database; the patient-clinician database was more varied with respect to sex and social class. Both databases were skewed towards Whites. Fairness in particular may have differed in a more heterogeneous database. 
Third, while both databases were comparable in size to those used in previous work, much larger numbers of participants would be preferable to assess the stability of the findings.  To meet the need for much larger databases is a pressing problem for research with clinical and other sensitive populations.  Visual behavior, as well as audio and speech, are personally identifiable modalities.  People with clinical disorders, such as depression, are reluctant to make their data broadly available.  Solutions to this problem are needed.


\section*{ETHICAL IMPACT STATEMENT}
Data were from two prior studies. Data collection protocols and data analysis for each were approved by the respective institutional Ethics Review Boards. All participants' gave informed consent prior to data collection. The algorithms we developed are intended for research use.  Clinical use would require further development and testing to establish their validity for the intended populations. 





{\small
\bibliographystyle{ieee}
\bibliography{egbib}

@article{alghowinem2020interpretation,
  title={Interpretation of depression detection models via feature selection methods},
  author={Alghowinem, Sharifa Mohammed and Gedeon, Tom and Goecke, Roland and Cohn, Jeffrey and Parker, Gordon},
  journal={IEEE Transactions on Affective Computing},
  year={2020},
  publisher={IEEE}
}

@article{prinz1979multivariate,
  title={Multivariate assessment of conflict in distressed and nondistressed mother-adolescent dyads},
  author={Prinz, Ronald J and Foster, Sharon and Kent, Ronald N and O'Leary, K Daniel},
  journal={Journal of applied behavior analysis},
  volume={12},
  number={4},
  pages={691--700},
  year={1979},
  publisher={Wiley Online Library}
}

@article{dibekliouglu2017dynamic,
  title={Dynamic multimodal measurement of depression severity using deep autoencoding},
  author={Dibeklio{\u{g}}lu, Hamdi and Hammal, Zakia and Cohn, Jeffrey F},
  journal={IEEE journal of biomedical and health informatics},
  volume={22},
  number={2},
  pages={525--536},
  year={2017},
  publisher={IEEE}
}

@article{girard2014nonverbal,
  title={Nonverbal social withdrawal in depression: Evidence from manual and automatic analyses},
  author={Girard, Jeffrey M and Cohn, Jeffrey F and Mahoor, Mohammad H and Mavadati, S Mohammad and Hammal, Zakia and Rosenwald, Dean P},
  journal={Image and vision computing},
  volume={32},
  number={10},
  pages={641--647},
  year={2014},
  publisher={Elsevier}
}

@article{yang2012detecting,
  title={Detecting depression severity from vocal prosody},
  author={Yang, Ying and Fairbairn, Catherine and Cohn, Jeffrey F},
  journal={IEEE transactions on affective computing},
  volume={4},
  number={2},
  pages={142--150},
  year={2012},
  publisher={IEEE}
}

@inproceedings{girard2017sayette,
  title={Sayette group formation task (GFT) spontaneous facial expression database},
  author={Girard, Jeffrey M and Chu, Wen-Sheng and Jeni, L{\'a}szl{\'o} A and Cohn, Jeffrey F},
  booktitle={FG},
  pages={581--588},
  year={2017},
  organization={IEEE}
}

@inproceedings{kacem2018detecting,
  title={Detecting depression severity by interpretable representations of motion dynamics},
  author={Kacem, Anis and Hammal, Zakia and Daoudi, Mohamed and Cohn, Jeffrey},
  booktitle={FG},
  pages={739--745},
  year={2018},
  organization={IEEE}
}

@article{kroenke2009phq,
  title={The PHQ-8 as a measure of current depression in the general population},
  author={Kroenke, Kurt and Strine, Tara W and Spitzer, Robert L and Williams, Janet BW and Berry, Joyce T and Mokdad, Ali H},
  journal={Journal of affective disorders},
  volume={114},
  number={1-3},
  pages={163--173},
  year={2009},
  publisher={Elsevier}
}

@article{nelson2021affective,
  title={Affective and Autonomic Reactivity During Parent--Child Interactions in Depressed and Non-Depressed Mothers and Their Adolescent Offspring},
  author={Nelson, Benjamin W and Sheeber, Lisa and Pfeifer, Jennifer H and Allen, Nicholas B},
  journal={Research on Child and Adolescent Psychopathology},
  volume={49},
  number={11},
  pages={1513--1526},
  year={2021},
  publisher={Springer}
}

@article{bilalpurmultimodal,
  title={Multimodal Feature Selection for Detecting Mothers’ Depression in Dyadic Interactions with their Adolescent Offspring},
  author={Bilalpur, Maneesh and Hinduja, Saurabh and Cariola, Laura A and Sheeber, Lisa B and Allen, Nick and Jeni, L{\'a}szl{\'o} A and Morency, Louis-Philippe and Cohn, Jeffrey F},
  journal={FG},
  year={2023},
  organization={IEEE}
}

@inproceedings{ringeval2019avec,
  title={AVEC 2019 workshop and challenge: state-of-mind, detecting depression with AI, and cross-cultural affect recognition},
  author={Ringeval, Fabien and Schuller, Bj{\"o}rn and Valstar, Michel and Cummins, Nicholas and Cowie, Roddy and Tavabi, Leili and Schmitt, Maximilian and Alisamir, Sina and Amiriparian, Shahin and Messner, Eva-Maria and others},
  booktitle={9th AVEC Challenge},
  pages={3--12},
  year={2019}
}

@article{song2020spectral,
  title={Spectral representation of behaviour primitives for depression analysis},
  author={Song, Siyang and Jaiswal, Shashank and Shen, Linlin and Valstar, Michel},
  journal={IEEE Transactions on Affective Computing},
  volume={13},
  number={2},
  pages={829--844},
  year={2022},
  publisher={IEEE}
}

@inproceedings{wu2023self,
  title={Self-supervised representations in speech-based depression detection},
  author={Wu, Wen and Zhang, Chao and Woodland, Philip C},
  booktitle={ICASSP 2023-2023 IEEE International Conference on Acoustics, Speech and Signal Processing (ICASSP)},
  pages={1--5},
  year={2023},
  organization={IEEE}
}

@inproceedings{daoudi2019gram,
  title={Gram matrices formulation of body shape motion: an application for depression severity assessment},
  author={Daoudi, Mohamed and Hammal, Zakia and Kacem, Anis and Cohn, Jeffrey F},
  booktitle={2019 8th International Conference on Affective Computing and Intelligent Interaction Workshops and Demos (ACIIW)},
  pages={258--263},
  year={2019},
  organization={IEEE}
}

@article{scherer2014automatic,
  title={Automatic audiovisual behavior descriptors for psychological disorder analysis},
  author={Scherer, Stefan and Stratou, Giota and Lucas, Gale and Mahmoud, Marwa and Boberg, Jill and Gratch, Jonathan and Morency, Louis-Philippe and others},
  journal={Image and Vision Computing},
  volume={32},
  number={10},
  pages={648--658},
  year={2014},
  publisher={Elsevier}
}

@inproceedings{bilalpur2023shap,
  title={SHAP-based Prediction of Mother's History of Depression to Understand the Influence on Child Behavior},
  author={Bilalpur, Maneesh and Hinduja, Saurabh and Cariola, Laura and Sheeber, Lisa and Allen, Nicholas and Morency, Louis-Philippe and Cohn, Jeffrey F},
  booktitle={Proceedings of the 25th International Conference on Multimodal Interaction},
  pages={537--544},
  year={2023}
}

@inproceedings{alghowinem2013eye,
  title={Eye movement analysis for depression detection},
  author={Alghowinem, Sharifa and Goecke, Roland and Wagner, Michael and Parker, Gordon and Breakspear, Michael},
  booktitle={2013 IEEE International Conference on Image Processing},
  pages={4220--4224},
  year={2013},
  organization={IEEE}
}

@article{hamilton1960rating,
  title={A rating scale for depression},
  author={Hamilton, Max},
  journal={Journal of neurology, neurosurgery, and psychiatry},
  volume={23},
  number={1},
  pages={56},
  year={1960},
  publisher={BMJ Publishing Group}
}

@article{al2018video,
  title={Video-based depression level analysis by encoding deep spatiotemporal features},
  author={Al Jazaery, Mohamad and Guo, Guodong},
  journal={IEEE Transactions on Affective Computing},
  volume={12},
  number={1},
  pages={262--268},
  year={2018},
  publisher={IEEE}
}

@article{greenberg2023economic,
  title={The economic burden of adults with major depressive disorder in the United States (2019)},
  author={Greenberg, Paul and Chitnis, Abhishek and Louie, Derek and Suthoff, Ellison and Chen, Shih-Yin and Maitland, Jessica and Gagnon-Sanschagrin, Patrick and Fournier, Andree-Anne and Kessler, Ronald C},
  journal={Advances in Therapy},
  volume={40},
  number={10},
  pages={4460--4479},
  year={2023},
  publisher={Springer}
}

@inproceedings{cohn2009detecting,
  title={Detecting depression from facial actions and vocal prosody},
  author={Cohn, Jeffrey F and Kruez, Tomas Simon and Matthews, Iain and Yang, Ying and Nguyen, Minh Hoai and Padilla, Margara Tejera and Zhou, Feng and De la Torre, Fernando},
  booktitle={2009 3rd international conference on affective computing and intelligent interaction and workshops},
  pages={1--7},
  year={2009},
  organization={IEEE}
}

@article{burcusa2007risk,
  title={Risk for recurrence in depression},
  author={Burcusa, Stephanie L and Iacono, William G},
  journal={Clinical psychology review},
  volume={27},
  number={8},
  pages={959--985},
  year={2007},
  publisher={Elsevier}
}

@article{ning2024representation,
  title={Representation learning and identity adversarial training for facial behavior understanding},
  author={Ning, Mang and Salah, Albert Ali and Ertugrul, Itir Onal},
  journal={arXiv preprint arXiv:2407.11243},
  year={2024}
}

@inproceedings{he2022masked,
  title={Masked autoencoders are scalable vision learners},
  author={He, Kaiming and Chen, Xinlei and Xie, Saining and Li, Yanghao and Doll{\'a}r, Piotr and Girshick, Ross},
  booktitle={Proceedings of the IEEE/CVF conference on computer vision and pattern recognition},
  pages={16000--16009},
  year={2022}
}

@article{adam2014method,
  title={A method for stochastic optimization},
  author={Adam, Kingma DP Ba J and others},
  journal={arXiv preprint arXiv:1412.6980},
  volume={1412},
  number={6},
  year={2014}
}

@article{hardt2016equality,
  title={Equality of opportunity in supervised learning},
  author={Hardt, Moritz and Price, Eric and Srebro, Nati},
  journal={Advances in neural information processing systems},
  volume={29},
  year={2016}
}

@inproceedings{ertugrul2019afar,
  title={AFAR: A Deep Learning Based Tool for Automated Facial Affect Recognition},
  author={Onal Ertugrul, Itir and Jeni, L{\'a}szl{\'o} A and Ding, Wanqiao and Cohn, Jeffrey F},
  booktitle={2019 14th IEEE International Conference on Automatic Face \& Gesture Recognition (FG 2019)},
  year={2019},
  organization={IEEE}
}

@article{nelson2012african,
  title={African American and European American mothers' beliefs about negative emotions and emotion socialization practices},
  author={Nelson, Jackie A and Leerkes, Esther M and O'Brien, Marion and Calkins, Susan D and Marcovitch, Stuart},
  journal={Parenting},
  volume={12},
  number={1},
  pages={22--41},
  year={2012},
  publisher={Taylor \& Francis}
}

@article{frank2011predictors,
  title={Predictors and moderators of time to remission of major depression with interpersonal psychotherapy and SSRI pharmacotherapy},
  author={Frank, Ellen and Cassano, Giovanni B and Rucci, Paola and Thompson, Wesley K and Kraemer, Helena C and Fagiolini, Andrea and Maggi, Luca and Kupfer, David J and Shear, M Katherine and Houck, Patricia R and others},
  journal={Psychological medicine},
  volume={41},
  number={1},
  pages={151--162},
  year={2011},
  publisher={Cambridge University Press}
}

@article{sheeber2023maternal,
  title={Maternal aggressive behavior in interactions with adolescent offspring: Proximal social--cognitive predictors in depressed and nondepressed mothers.},
  author={Sheeber, Lisa and Lougheed, Jessica and Hollenstein, Tom and Leve, Craig and Mudiam, Kavya and Diercks, Catherine and Allen, Nicholas},
  journal={Journal of psychopathology and clinical science},
  volume={132},
  number={8},
  pages={1019},
  year={2023},
  publisher={American Psychological Association}
}

@inproceedings{cheong2023towards,
  title={Towards gender fairness for mental health prediction},
  author={Cheong, Jiaee and Kuzucu, Selim and Kalkan, Sinan and Gunes, Hatice},
  year={2023},
  journal={International Joint Conferences on Artificial Intelligence Organization}
}

@inproceedings{cheong2024fairrefuse,
  title={Fairrefuse: Referee-guided fusion for multi-modal causal fairness in depression detection},
  author={Cheong, Jiaee and Kalkan, Sinan and Gunes, Hatice},
  booktitle={International Joint Conference on Artificial Intelligence (IJCAI)},
  year={2024}
}

@inproceedings{gratch2014distress,
  title={The distress analysis interview corpus of human and computer interviews.},
  author={Gratch, Jonathan and Artstein, Ron and Lucas, Gale M and Stratou, Giota and Scherer, Stefan and Nazarian, Angela and Wood, Rachel and Boberg, Jill and DeVault, David and Marsella, Stacy and others},
  booktitle={Lrec},
  volume={14},
  pages={3123--3128},
  year={2014},
  organization={Reykjavik}
}

@article{arioz2022scoping,
  title={Scoping review on the multimodal classification of depression and experimental study on existing multimodal models},
  author={Arioz, Umut and Smrke, Ur{\v{s}}ka and Plohl, Nejc and Mlakar, Izidor},
  journal={Diagnostics},
  volume={12},
  number={11},
  pages={2683},
  year={2022},
  publisher={MDPI}
}

@inproceedings{walmer2023teaching,
  title={Teaching matters: Investigating the role of supervision in vision transformers},
  author={Walmer, Matthew and Suri, Saksham and Gupta, Kamal and Shrivastava, Abhinav},
  booktitle={Proceedings of the IEEE/CVF Conference on Computer Vision and Pattern Recognition},
  pages={7486--7496},
  year={2023}
}

@inproceedings{koutlis2024leveraging,
  title={Leveraging representations from intermediate encoder-blocks for synthetic image detection},
  author={Koutlis, Christos and Papadopoulos, Symeon},
  booktitle={European Conference on Computer Vision},
  pages={394--411},
  year={2024},
  organization={Springer}
}

@inproceedings{alghowinem2015cross,
  title={Cross-cultural detection of depression from nonverbal behaviour},
  author={Alghowinem, Sharifa and Goecke, Roland and Cohn, Jeffrey F and Wagner, Michael and Parker, Gordon and Breakspear, Michael},
  booktitle={2015 11th IEEE International conference and workshops on automatic face and gesture recognition (FG)},
  volume={1},
  pages={1--8},
  year={2015},
  organization={IEEE}
}

@article{dehdashtian2024fairness,
  title={Fairness and Bias Mitigation in Computer Vision: A Survey},
  author={Dehdashtian, Sepehr and He, Ruozhen and Li, Yi and Balakrishnan, Guha and Vasconcelos, Nuno and Ordonez, Vicente and Boddeti, Vishnu Naresh},
  journal={arXiv preprint arXiv:2408.02464},
  year={2024}
}

@article{ranjit2023variation,
  title={Variation of gender biases in visual recognition models before and after finetuning},
  author={Ranjit, Jaspreet and Wang, Tianlu and Ray, Baishakhi and Ordonez, Vicente},
  journal={arXiv preprint arXiv:2303.07615},
  year={2023}
}

@inproceedings{zhang2016yin,
  title={Yin and yang: Balancing and answering binary visual questions},
  author={Zhang, Peng and Goyal, Yash and Summers-Stay, Douglas and Batra, Dhruv and Parikh, Devi},
  booktitle={Proceedings of the IEEE conference on computer vision and pattern recognition},
  pages={5014--5022},
  year={2016}
}

@inproceedings{birhane2024dark,
  title={The dark side of dataset scaling: Evaluating racial classification in multimodal models},
  author={Birhane, Abeba and Dehdashtian, Sepehr and Prabhu, Vinay and Boddeti, Vishnu},
  booktitle={Proceedings of the 2024 ACM Conference on Fairness, Accountability, and Transparency},
  pages={1229--1244},
  year={2024}
}

@inproceedings{zhao2021understanding,
  title={Understanding and evaluating racial biases in image captioning},
  author={Zhao, Dora and Wang, Angelina and Russakovsky, Olga},
  booktitle={Proceedings of the IEEE/CVF international conference on computer vision},
  pages={14830--14840},
  year={2021}
}

@inproceedings{hirota2022quantifying,
  title={Quantifying societal bias amplification in image captioning},
  author={Hirota, Yusuke and Nakashima, Yuta and Garcia, Noa},
  booktitle={Proceedings of the IEEE/CVF conference on computer vision and pattern recognition},
  pages={13450--13459},
  year={2022}
}

@inproceedings{fu2025first,
  title={The first MPDD challenge: multimodal personality-aware depression detection},
  author={Fu, Changzeng and Fu, Zelin and Zhang, Qi and Kuang, Xinhe and Dong, Jiacheng and Su, Kaifeng and Su, Yikai and Shi, Wenbo and Yao, Junfeng and Zhao, Yuliang and others},
  booktitle={Proceedings of the 33rd ACM International Conference on Multimedia},
  pages={13924--13929},
  year={2025}
}

@inproceedings{ilias2024cross,
  title={A cross-attention layer coupled with multimodal fusion methods for recognizing depression from spontaneous speech},
  author={Ilias, Loukas and Askounis, Dimitris},
  booktitle={Proc. Interspeech},
  volume={2024},
  pages={912--916},
  year={2024}
}
}

\end{document}